\documentclass[letterpaper]{article} 
\usepackage{aaai2026}  
\usepackage{times}  
\usepackage{helvet}  
\usepackage{courier}  
\usepackage[hyphens]{url}  
\usepackage{graphicx} 
\usepackage{natbib}  
\usepackage{caption} 
\usepackage{algorithm}
\usepackage{algorithmic}

\usepackage{enumitem}
\usepackage{amsmath}
\usepackage{amsfonts}
\usepackage{booktabs}
\usepackage[capitalize]{cleveref}
\usepackage{nameref}
\usepackage{float}

\crefname{section}{Sec.}{Secs.}
\crefname{section}{Section}{Sections}
\crefname{table}{Table}{Tables}
\crefname{table}{Tab.}{Tabs.}

\usepackage{multirow}
\usepackage[table]{xcolor}

\def\model{SRAM}
\def\ds{RealismGrading}

\makeatletter

\newcommand{\Rmnum}[1]{\textcolor{red}{\expandafter\@slowromancap\romannumeral #1@}}

\makeatother

\usepackage{newfloat}
\usepackage{listings}
\DeclareCaptionStyle{ruled}{labelfont=normalfont,labelsep=colon,strut=off} 
\floatstyle{ruled}
\newfloat{listing}{tb}{lst}{}
\floatname{listing}{Listing}
\title{SRAM: Shape-Realism Alignment Metric for No Reference 3D Shape Evaluation}
\author{
    Sheng Liu, Tianyu Luan\thanks{Corresponding author}, Phani Nuney, Xuelu Feng, Junsong Yuan
}
\affiliations{
    State University of New York at Buffalo\\
    \{sliu66, tianyulu, phaneesw, xuelufen, jsyuan\} @buffalo.edu
}

\usepackage{bibentry}
\begin{document}

\maketitle

\begin{abstract}
3D generation and reconstruction techniques have been widely used in computer games, film, and other content creation areas. As the application grows, there is a growing demand for 3D shapes that look truly realistic. Traditional evaluation methods rely on a ground truth to measure mesh fidelity. However, in many practical cases, a shape's realism does not depend on having a ground truth reference. In this work, we propose a Shape-Realism Alignment Metric that leverages a large language model (LLM) as a bridge between mesh shape information and realism evaluation. To achieve this, we adopt a mesh encoding approach that converts 3D shapes into the language token space. A dedicated realism decoder is designed to align the language model’s output with human perception of realism. Additionally, we introduce a new dataset, \ds{}, which provides human-annotated realism scores without the need for ground truth shapes.  Our dataset includes shapes generated by 16 different algorithms on over a dozen objects, making it more representative of practical 3D shape distributions. We validate our metric's performance and generalizability through k-fold cross-validation across different objects. Experimental results show that our metric correlates well with human perceptions and outperforms existing methods, and has good generalizability.
\end{abstract}

\begin{figure*}
  \centering
  \includegraphics[width=0.8\linewidth]{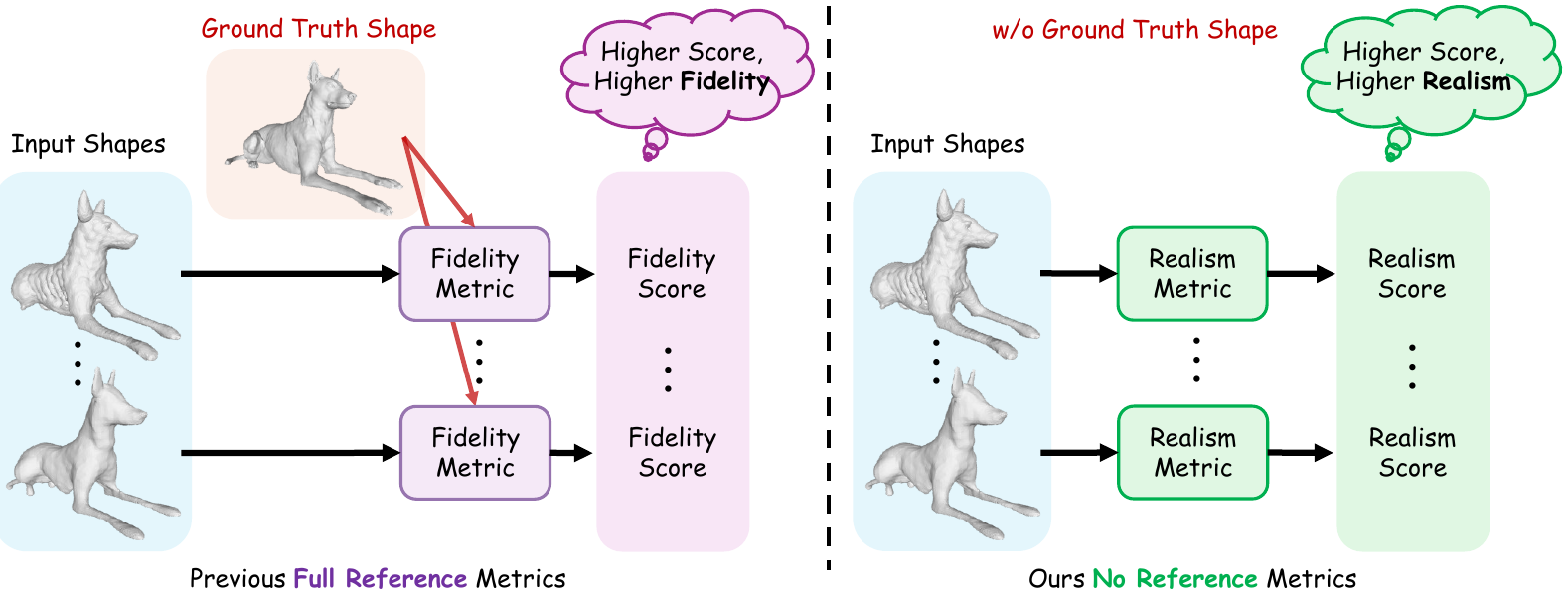}
  \caption{Full reference 3D shape evaluation vs. no reference 3D shape evaluation. Left: Traditional metrics require ground truth references to evaluate the fidelity of 3D shapes. Right: Our metric can evaluate 3D shape realism without a reference. In this paper, we refer ``no reference fidelity'' as ``realism'' since fidelity is typically based on comparisons while realism is not.}
  \label{fig:teaser}
\end{figure*}

\section{Introduction}

3D generation and reconstruction technologies are widely used in computer games, film and television production, and many emerging content creation applications. There is an increasing demand for 3D shapes that appear genuinely realistic to the human eye. Therefore, it is crucial to develop an evaluation metric aligned with human perception to assess the realism of 3D shapes, guiding the creation of more lifelike 3D shapes and contents.

Previous works, such as \cite{luan2024spectrum}, have mainly focused on evaluating mesh fidelity. As shown in \cref{fig:teaser}, these evaluations (shown on the left) require a ground truth as a reference. However, in many practical applications, a shape’s realism does not necessarily depend on a ground truth. For instance, people typically consider a reconstructed animal as lacking realism. In many 3D generation applications, assessing the realism of a generated shape without relying on a ground truth reference is a common challenge, yet previous works have not adequately addressed this issue.

To overcome the dependence on ground truth for mesh realism evaluation, we design a metric that takes only the 3D shape as input. As shown in \cref{fig:teaser} right half, although realism can exist independently of a ground truth, capturing such information without it is challenging. Full-reference metrics, such as \cite{luan2024spectrum}, measure fidelity by comparing the input shape with a ground truth. In contrast, a no-reference realism metric needs to understand the shape by itself at a detailed and semantic level. In full-reference approaches, the magnitude of the difference between two shapes is inversely related to fidelity, making the mapping more regular and easier to train. However, in no-reference metrics, the scale (or magnitude) of a mesh does not explicitly correlate with its realism, so the metric must instead capture the high-level semantic properties of the shape. Here, we refer ``no reference fidelity'' as ``realism'' since fidelity is typically based on comparisons while realism is not.

We propose to use a large language model (LLM) as a bridge to provide alignment between mesh shape and realism. LLMs have rich high-level knowledge priors and reasoning capabilities; a well-trained language model can provide the reasoning and knowledge priors needed for our metric to map from shape to realism. If we can align 3D shape information at the input side to the language model, and map the language model's output to realism at the output side, we can achieve better alignment from shape to realism. Through this alignment, we can evaluate shape realism without needing a ground truth shape reference.

We name our proposed metric Shape-Realism Alignment Metric (\model{}). To align mesh shape information with the language model input, we adopt the mesh encoding approach from \cite{pointbert} to encode mesh shapes into the language token space. We also design a realism decoder to align the language model's output with human perception of realism. Additionally, to provide the network with training data that includes human-annotated realism scores, we introduce a human-annotated dataset named \ds{}. Unlike previous datasets such as \cite{nehme2023textured}, our dataset's scoring is based on the shape itself without requiring ground truth references. More importantly, the meshes in our dataset are based on real reconstruction and generation algorithms rather than synthetic ones, making this dataset better reflect the distribution of shapes in real-world 3D reconstruction and generation. \ds{} comprises over a dozen objects, and for each object, we have 9-16 distortion shapes generated by a set of real-world reconstruction and generation algorithms. Furthermore, we obtained realism annotations for these distortion shapes from hundreds of human subjects. In our experiments, we validate our metric's performance and generalizability through k-fold cross-validation across different objects.

In summary, our contributions are as follows:
\vspace{-1mm}
\begin{itemize}
    \item We propose a metric for evaluating shape realism that does not require a ground truth mesh as a reference, enabling a no-reference assessment of a mesh's realism.
    \item We design a language-model-based Shape-Realism alignment pipeline. This pipeline can leverage the language model as a bridge to align the shape feature space with realism feature space, so that the metric can evaluate realism based only on 3D mesh shape without a ground truth mesh shape reference.
    \item We introduce the \ds{} dataset, which provides human-annotated training data for our no-reference metric. In our dataset, annotations focus solely on the shape itself, without any comparison to a ground truth. Additionally, the meshes are produced by real reconstruction and generation algorithms rather than being synthetic.
\end{itemize}

Our experiments demonstrate that our \model{} metric correlates more closely with human perceptions of realism than existing state-of-the-art metrics. 

\section{Related Work}
\textbf{3D shape metrics commonly used in 3D generation and reconstruction.}
Evaluating 3D shape quality remains challenging due to a mismatch between geometric metrics and human perception. Euclidean distance such as Chamfer Distance (CD)\cite{borgefors1984CD, luan2021pc, zhai2023language, luan2023high, gong2022self,  zhao2025pp, luan2024divide, luan2025scalable, gong2023progressive, wu2024fsc, song2022pref, zhang2021learning}, IoU\cite{hu2021iou, chen2021iou, nie2020totaliou, henderson2018iou, tang2022iou, santhanam2023iou}, F-score~\cite{wang2018fscore, genova2020fscore, bechtold2021fscore, tatarchenko2019fscore}, and UHD~\cite{wu2020uhd} emphasize geometry but miss semantic and structural cues. For generative models, distribution-level metrics like MMD~\cite{achlioptas2018mmd}, JSD~\cite{kullback1951jsd}, TMD~\cite{wu2020uhd}, SCEU~\cite{wang2022groupdancer, wang2024dancecamanimator}, and FPD~\cite{shu2019fpd} assess set-level quality, not individual fidelity. A parallel line of research has focused on evaluating shape quality in the context of 3D mesh compression and watermarking applications. Works such as~\cite{wang2010wm, bulbul2011wm, lavoue2009wm, corsini2013wm}. Our work addresses this gap by introducing a perceptually aligned metric for evaluating single 3D shapes.

\textbf{Perceptual quality assessment for 3D shapes.}
To overcome the limitations of geometric metrics, recent works~\cite{sarvestani2024hybridmqa, yang2023tsmd, yang2024tdmd, cui2024sjtu, nehme2020visual, nehme2023textured} propose perceptual quality metrics using deep learning. HybridMQA~\cite{sarvestani2024hybridmqa} fuses multiple modalities, while TSMD~\cite{yang2023tsmd}, TDMD~\cite{yang2024tdmd}, and SJTU-PQA~\cite{cui2024sjtu} incorporate texture cues for textured mesh evaluation. However, these full-reference methods rely on ground truth and focus on surface appearance, limiting applicability to textureless or real-world data. Moreover, they are trained on synthetic distortions, unlike real-world artifacts. In contrast, our method requires no reference, targets geometry-only evaluation, and learns from authentic distortions in practical 3D tasks.

\begin{figure*}
  \centering
  \includegraphics[width=0.79\linewidth]{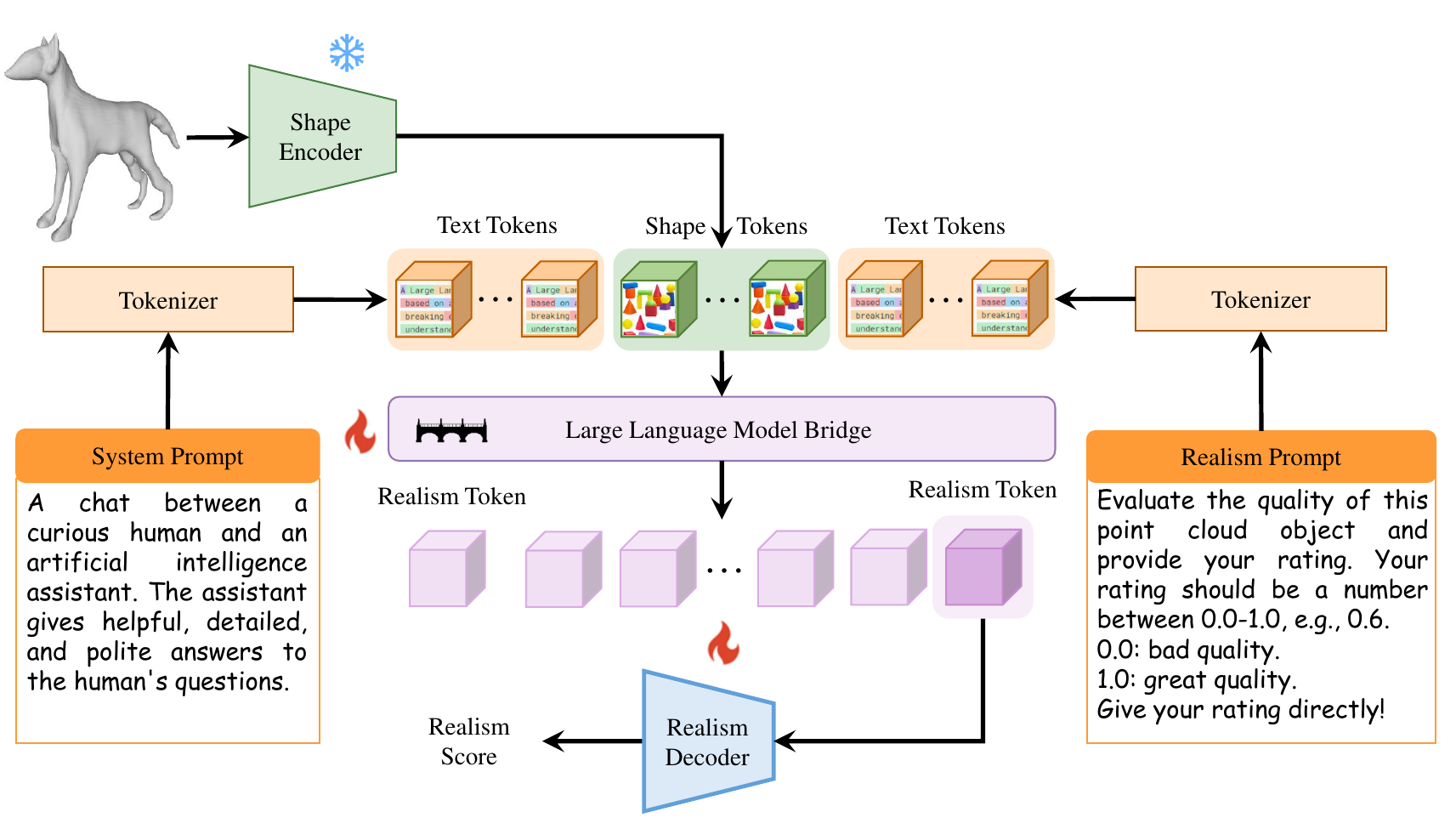}
  \vspace{-2pt}
  \caption{The pipeline of our Shape-Realism Alignment Metric (\model{}). Our metric can take a mesh shape as input and measure its realism without a ground truth mesh shape reference. It uses a language model as a bridge to achieve alignment from 3D shape to realism score. The language model bridge has 3 inputs: text tokens from the system prompt, 3D shape tokens from the 3D shape encoder, and another part of text tokens from the realism prompt. In the output part of our model, we design a token-based realism decoder to align language tokens with realism scores.}
  \label{fig:pipeline}
\end{figure*}

\section{Method}

\subsection{Problem Formulation}

We formulate the problem of defining a realism metric as follows. Given an input mesh $x$, we design a metric function $F(\cdot)$, whose output is the realism corresponding to the mesh, i.e.,
\begin{equation}
    s = F(x;\theta),
\end{equation}
where $F(x;\theta)$ is a neural network, and $\theta$ is the learnable parameters of the network. 

In the following sections, we will introduce how we design the metric $F(x;\theta)$ and how we train it. Overall, our training process can be expressed as:
\begin{equation}
    \min_\theta{\mathcal{L}(F(x;\theta) - y_x)},
\end{equation}
where $y$ is the human annotation of mesh shape input $x$.

\subsection{Shape-Realism Alignment Metric}
\label{sec:sram}

We show our pipeline design in \cref{fig:pipeline}. Our metric can take a mesh shape as input and measure its realism without a ground truth mesh shape reference. It uses an LLM as a bridge to achieve alignment from 3D shape to realism score. Specifically, the input of the language bridge consists of three parts. The first part is the system prompt, which establishes a context for the language model. The second part is the shape tokens, where we use a PointNet-based 3D shape encoder to encode a 3D mesh into tokens. The third part is the realism prompt. This input helps elicit our language model's ability to reason about realism of meshes.

In the output part of our model, we use a token-based realism decoder to regress realism scores. This way, we achieve alignment from mesh shape to realism score, thereby evaluating mesh shape realism without depending on ground truth mesh references. In the remainder of this subsection, we introduce the design of each module of our Shape-Realism Alignment Metric (\model{}).

\textbf{System prompt.}
In the context of LLM, the system prompt primarily provides high-level context to the pretrained language model. It may also constrain and guide the language model's behavior. In our metric design, we set the system prompt as a user/assistant Q\&A model. We follow \cite{xu2024pointllm}'s system prompt to prepare the language model system with overall contextual information.

\textbf{Shape encoder.}
The shape encoder extracts shape features. Considering the diversity of our input shapes and the limitations of our human-annotated 3D shape realism data in terms of shape variety, we need a 3D shape encoding model with good context to meet our generalization requirements for shape variety. Therefore, we consider using a large shape dataset pretrained model for this module. We follow the design in Point-BERT \cite{yu2022point}. Here, we treat the vertex of the input mesh shape as a point cloud. Point-BERT's structure can establish good alignment from point clouds to text tokens through the approach of point cloud masking. Point-BERT provides pretrained models on ScanObjectNN \cite{uy-scanobjectnn-iccv19} and ShapeNet \cite{chang2015shapenet}. We can leverage the shape priors inherent in these pretrained models to benefit our shape realism evaluation.

\begin{table*}[ht]
    \centering
    \resizebox{\linewidth}{!}{
    \begin{tabular}{l|cccccccc}
    \hline
         Object Type & CRM~\cite{wang2024crm} & DreamGaussian~\cite{tang2023dreamgaussian} & InstantMesh~\cite{xu2024instantmesh} & LGM~\cite{tang2024lgm} & One2345~\cite{liu2023one} & One2345pp~\cite{liu2024one} & PeRFlowText~\cite{yan2024perflow} & ShapE~\cite{jun2023shap}  \\ 
    \hline
        01-Dog~\cite{renderbot} & \checkmark & \checkmark & \checkmark & \checkmark & \checkmark & \checkmark & \checkmark & \checkmark  \\ 
        02-Fish~\cite{arc3d} & \checkmark & \checkmark & \checkmark & \checkmark & \checkmark & \checkmark & \checkmark & \checkmark  \\ 
        03-Female Face~\cite{zhu2023facescape,yang2020facescape} & \checkmark & \checkmark & \checkmark & \checkmark & \checkmark & ~ & \checkmark & ~  \\ 
        04-Male Face~\cite{wang2022faceverse} & \checkmark & \checkmark & \checkmark & \checkmark & \checkmark & ~ & \checkmark & ~  \\ 
        05-Female Human Body A~\cite{yu2021function4d} & \checkmark & \checkmark & \checkmark & \checkmark & ~ & \checkmark & \checkmark & \checkmark  \\ 
        06-Female Human Body B~\cite{yu2021function4d} & \checkmark & \checkmark & \checkmark & \checkmark & ~ & \checkmark & \checkmark & \checkmark  \\ 
        07-Hand w/ Arm~\cite{arc3d} & \checkmark & \checkmark & \checkmark & \checkmark & \checkmark & ~ & \checkmark & \checkmark  \\ 
        08-Hand w/o Arm~\cite{arc3d} & \checkmark & \checkmark & \checkmark & \checkmark & \checkmark & ~ & \checkmark & \checkmark  \\
        09-Male Human Body A~\cite{yu2021function4d} & \checkmark & \checkmark & \checkmark & \checkmark & ~ & \checkmark & \checkmark & \checkmark  \\ 
        10-Male Human Body B~\cite{yu2021function4d} & \checkmark & \checkmark & \checkmark & \checkmark & ~ & \checkmark & \checkmark & \checkmark  \\
        11-Keyboard~\cite{downs2022google} & \checkmark & \checkmark & \checkmark & \checkmark & ~ & \checkmark & \checkmark & \checkmark  \\ 
        12-Building & \checkmark & \checkmark & \checkmark & ~ & ~ & \checkmark & \checkmark & \checkmark  \\ 
        13-Bus~\cite{arc3d} & \checkmark & \checkmark & \checkmark & \checkmark & ~ & \checkmark & \checkmark & \checkmark  \\ 
        14-Mug~\cite{downs2022google} & \checkmark & \checkmark & \checkmark & \checkmark & \checkmark & \checkmark & \checkmark & \checkmark \\ 
        15-Plant~\cite{sketchfab} & \checkmark & \checkmark & \checkmark & \checkmark & ~ & \checkmark & \checkmark & \checkmark \\ 
        16-Shoe~\cite{downs2022google} & \checkmark & \checkmark & \checkmark & \checkmark & \checkmark & \checkmark & \checkmark & \checkmark \\ 
        
    \hline
    \end{tabular}
    }
    \caption{A list of the dataset and the 3D reconstruction/generation method we used. $\checkmark$ indicates that the corresponding method was employed to generate the mesh for that object category. (Part 1)}
    \label{tab:dataset1}
    \vspace{-2pt}
\end{table*}

\begin{table*}[ht]
    \centering
    \resizebox{\linewidth}{!}{
    \begin{tabular}{l|cccccccc}
    \hline
         Object Type & ShapEText~\cite{jun2023shap} & SplatterImage~\cite{szymanowicz24splatter} & TripoSR~\cite{TripoSR2024} & ECON~\cite{xiu2023econ} & ICON~\cite{xiu2022icon} & PiFU~\cite{saito2019pifu} & PiFUHD~\cite{saito2020pifuhd} & HaMeR~\cite{pavlakos2024reconstructing} \\ 
    \hline
        01-Dog~\cite{renderbot} & \checkmark & \checkmark & \checkmark & ~ & ~ & ~ & ~ & ~  \\ 
        02-Fish~\cite{arc3d} & \checkmark & \checkmark & \checkmark & ~ & ~ & ~ & ~ & ~  \\ 
        03-Female Face~\cite{zhu2023facescape,yang2020facescape} & \checkmark & \checkmark & \checkmark & ~ & ~ & ~ & ~ & ~  \\ 
        04-Male Face~\cite{wang2022faceverse} & \checkmark & \checkmark & \checkmark & ~ & ~ & ~ & ~ & ~  \\ 
        05-Female Human Body A~\cite{yu2021function4d} & \checkmark & \checkmark & \checkmark & \checkmark & \checkmark & \checkmark & \checkmark & ~  \\ 
        06-Female Human Body B~\cite{yu2021function4d} & \checkmark & \checkmark & \checkmark & \checkmark & \checkmark & \checkmark & \checkmark & ~  \\ 
        07-Hand w/ Arm~\cite{arc3d} & \checkmark & ~ & \checkmark & ~ & ~ & ~ & ~ & \checkmark  \\ 
        08-Hand w/o Arm~\cite{arc3d} & \checkmark & ~ & \checkmark & ~ & ~ & ~ & ~ & \checkmark  \\
        09-Male Human Body A~\cite{yu2021function4d} & \checkmark & \checkmark & \checkmark & \checkmark & \checkmark & \checkmark & \checkmark & ~  \\ 
        10-Male Human Body B~\cite{yu2021function4d} & \checkmark & \checkmark & \checkmark & \checkmark & \checkmark & \checkmark & \checkmark & ~  \\
        11-Building & \checkmark & \checkmark & \checkmark & ~ & ~ & ~ & ~ & ~  \\ 
        12-Bus~\cite{arc3d} & \checkmark & ~ & \checkmark & ~ & ~ & ~ & ~ & ~  \\ 
        13-Keyboard~\cite{downs2022google}  & \checkmark & \checkmark & \checkmark & ~ & ~ & ~ & ~ & ~ \\ 
        14-Mug~\cite{downs2022google} & \checkmark & ~ & \checkmark & ~ & ~ & ~ & ~ & ~  \\ 
        15-Plant~\cite{sketchfab} & \checkmark & \checkmark & \checkmark & ~ & ~ & ~ & ~ & ~ \\ 
        16-Shoe~\cite{downs2022google} & \checkmark & \checkmark & \checkmark & ~ & ~ & ~ & ~ & ~ \\ 
    \hline
    \end{tabular}
    }
    \caption{A list of the dataset and the 3D reconstruction/generation method we used. $\checkmark$ indicates that the corresponding method was employed to generate the mesh for that object category. (Part 2)}
    \label{tab:dataset2}
    \vspace{-2pt}
\end{table*}

\textbf{Realism prompt.}
The realism prompt is the core prompt for using LLM for realism evaluation. Here, we need to design a prompt that can better elicit the LLM's abilities and prior knowledge to benefit our shape evaluation. We experiment with multiple possible prompts to find the most suitable one for our task. Notably, the realism prompt only provides an initial LLM prior for our metric. Our final realism evaluation does not completely rely on this prompt. During the training phase, we also rely on finetuning to optimize our model, enabling it to better learn the human realism annotations provided in our dataset, thereby better evaluating shape realism.

\textbf{Large language model bridge.}
The LLM bridge is a critical part in aligning shape information to realism information. Our model's high-level knowledge and reasoning abilities primarily depend on this part. We choose to initialize our LLM bridge with corresponding components from PointLLM \cite{xu2024pointllm}, a strong 3D LLM pre-trained using hundreds of thousands of 3D objects. The 3D LLM learns to understand semantics and details of 3D shapes during its training process. Such capability is crucial for realism evaluation. Our metric can utilize these capabilities along with appropriate finetuning to achieve better alignment between shape and realism.

\textbf{Realism decoder.}
The realism decoder is another key element in our metric design. We need to align text tokens with realism scores using this decoder. Considering the LLM's strong expressive capability and excellent pretraining, and the fact that our output score is only one-dimensional, we can incorporate a major part of the decoder function through fine-tuning the LLM itself. This design allows the LLM's output to already have a good realism representation, rather than placing the main alignment functionality on the decoder. This design can save the number of parameters of the decoder and better utilize the LLM's pretraining. Therefore, we are able to design our decoder in a simple form of an MLP. We aim to only summarize the realism score through the decoder and backpropagate regression gradients through the decoder during training, working with the LLM finetuning to achieve better text-to-realism alignment. Our experiments show that our simplified design can achieve good alignment with human perception.

\subsection{Training Process}

Based on the pre-trained shape encoder and LLM bridge, we design an end-to-end training procedure. First, to leverage the prior knowledge on shapes and language in the shape encoder and LLM bridge, we initially adopt the pre-training model from \cite{pointbert} and \cite{llama} for shape and language, respectively. Then, we randomly initialize the realism decoder. During training, the shape encoder is frozen while we simultaneously train the realism encoder and finetune the LLM bridge. This training strategy allows for better adaptation between the LLM and the realism decoder, enabling alignment from language tokens to realism scores through the light-weighted decoder.

\textbf{Training Loss.} Unlike the traditional LLM token classification loss, we employ a $l_2$ regression loss for the realism decoder. This design allows the network to learn continuous realism scoring information from human annotations. Our loss function is defined as follows:
\begin{equation}
    \mathcal{L} = \|F(x) - y_x\|_2,    
\end{equation}
where $x$ is the input mesh shape, $y_x$ is the realism human annotation of input $x$, $F(\cdot)$ is our metric network, and $\|\cdot\|$ is the $l_2$ loss.

\begin{table*}[htbp]
    
    \centering
    \resizebox{0.9\linewidth}{!}{
    \begin{tabular}{l|c|cccccccccccccccc|c}
    \hline 
    \multicolumn{2}{c|}{Object Number} & 1 & 2 & 3 & 4 & 5 & 6 & 7 & 8 & 9 & 10 & 11 & 12 & 13 & 14 & 15 & 16 &  Overall\tabularnewline
    \hline
    \multirow{2}{*}{PLCC}& NR-3DQA \cite{zhang2022no} & 0.47 & 0.70 & 0.59 & 0.70 & 0.18 & 0.51 & 0.60 & 0.76 & 0.41 & 0.51 & 0.82 & 0.55 & 0.61 & 0.41 & 0.34 & 0.74 & 0.555 \\
    & Ours & 0.87 & 0.31 & 0.88 & 0.65 & 0.9 & 0.94 & 0.39 & 0.69 & 0.82 & 0.85 & 0.76 & 0.38 & 0.46 & 0.67 & 0.81 & 0.64 & \textbf{0.689} \\
    \hline
     \multirow{2}{*}{SROCC}& Point-based& 0.48 & 0.51 & 0.87 & 0.57 & 0.35 & 0.62 & 0.47 & 0.43 & 0.56 & 0.38 & 0.95 & 0.52 & 0.84 & 0.75 & 0.49 & 0.68 & 0.591 \\
    & Ours & 0.92 & 0.44 & 0.88 & 0.61 & 0.74 & 0.88 & 0.39 & 0.7 & 0.85 & 0.9 & 0.89 & 0.62 & 0.25 & 0.51 & 0.9 & 0.66 & \textbf{0.696} \\
    \hline
    \multirow{2}{*}{KROCC}& Point-based & 0.31 & 0.40 & 0.72 & 0.39 & 0.37 & 0.44 & 0.35 & 0.31 & 0.42 & 0.18 & 0.85 & 0.39 & 0.71 & 0.56 & 0.33 & 0.53 & 0.454 \\
     & Ours & 0.78 & 0.37 & 0.76 & 0.46 & 0.57 & 0.73 & 0.33 & 0.6 & 0.67 & 0.72 & 0.7 & 0.56 & 0.13 & 0.4 & 0.73 & 0.54 & \textbf{0.565} \\
    \hline 
    \end{tabular}
    }
    \caption{We quantitatively compare our metric with an ad-hoc baseline using 3 correlation evaluations: PLCC \cite{pearson1920plcc}, SROCC \cite{spearman1910srocc}, and KROCC \cite{kendall1938new}, which are defined in Sec. \nameref{sec:eval}. Number 1-16 indicates the object index (aligns with object number in \cref{tab:dataset1} and \cref{tab:dataset2}). Bold numbers indicate the best performance.}
    \label{tab:sota}
\end{table*}

\begin{table}

\centering
\vspace{-2pt}
\resizebox{1\linewidth}{!}{
    \begin{tabular}{l|c|c}
    \hline
    Datasets & ShapeGrading \cite{luan2024spectrum} & \ds{} \\
    \hline
    Number of objects & 12 & 16\\ 
    Mesh shape types & Synthetic & Real-world \\
    Number of shape distortion types & 7 & 16 \\
    Needs ground truth for annotation & Yes & No \\
    Scoring range & $[0, 6]$ & $[0, 1]$ \\
    Scoring $95\%$ confidence interval & 0.303 & 0.077\\
    \hline
    \end{tabular}
}
\caption{We compare our dataset with ShapeGrading \cite{luan2024spectrum}. Our datasets do not use ground truth reference for human annotation, and the mesh shape in our dataset comes from real-world 3D reconstruction and generation methods.}
\label{tab:dataset}
\end{table}

\section{\ds{} Dataset Design}

We present a human-annotated dataset designed to capture the distortions introduced by real-world 3D reconstruction and generative methods. Our \ds{} dataset includes human annotations that evaluate the mesh shapes without ground truth reference, and it consists of meshes from real-world 3D reconstruction and generation methods. Our benchmark dataset construction involves three main stages: data generation, annotation, and evaluation. 

\subsection{Data Generation}
We begin by selecting 16 image/object pairs sourced from various datasets \cite{yu2021function4d,wang2022faceverse,zhu2023facescape,yang2020facescape,arc3d,renderbot,downs2022google,sketchfab,cgtrader}. For each pair, the input image is processed through 16 different algorithms (e.g., \cite{wang2024crm,tang2023dreamgaussian,xu2024instantmesh,tang2024lgm,liu2023one,liu2024one,yan2024perflow,jun2023shap,szymanowicz24splatter,TripoSR2024,xiu2023econ,xiu2022icon,saito2019pifu,saito2020pifuhd,pavlakos2024reconstructing}) to create distorted 3D meshes, while the associated object acts as the ground truth. For text-driven models, we first generate descriptive prompts using an image captioning technique and then use these prompts to produce the meshes. In \cref{tab:dataset1} and \cref{tab:dataset2}, we provide a complete list of the dataset and 3D reconstruction/generation method we used.

\subsection{Annotation Process} To obtain reliable distortion scores, we adopt a pairwise comparison method inspired by \cite{ponomarenko2009tid2008} and \cite{luan2024spectrum}. Each human evaluator compares 9-16 meshes of a specific object category with a given mesh material through 6 rounds of Swiss-system pairwise comparisons. Specifically, for each round, if the evaluator believes A mesh is more realistic than B mesh, then A mesh scores 1 in this round, and B mesh scores 0 in this round. Note that, in this process, the ground truth mesh is not shown to the human evaluators. After all 6 rounds, a mesh earns a score from 0 (losing every comparison) to 6 (winning every comparison). We gathered ratings from 319 subjects and got a total of 5,223 scores. These raw scores are then normalized to a $[0, 1]$ range. The overall reliability of the dataset is confirmed by an average 95\% confidence interval of approximately 5\%, computed with $\sigma_{\bar{x}} = z_{0.95}\sigma/\sqrt{N}$ (with $z_{0.95}\approx1.96$).

\begin{figure*}[t]
    \centering
    \includegraphics[trim={65pt 0 65pt 0},width=0.82\linewidth]{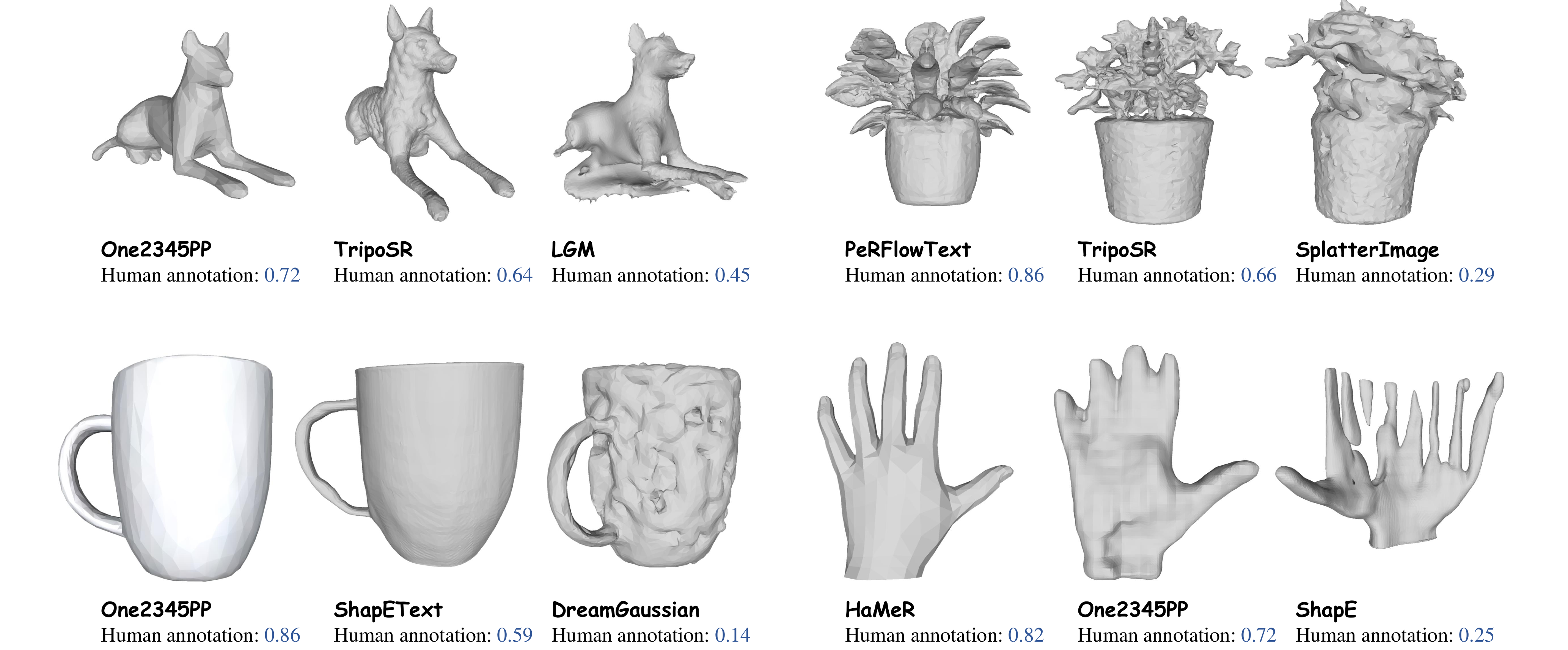}
    \caption{We show example meshes along with their human-annotated realism score from our \ds{} dataset. Methods used to produce these meshes are shown as well, \textit{e.g.}, ``\texttt{One2345pp}''. We observe that as the realism of a mesh increases, its annotated realism score also goes up.}
    \label{fig:vis_ds}
\end{figure*}

In \cref{tab:dataset}, we compare our dataset with the synthetic dataset presented in \cite{luan2024spectrum}. Our datasets do not use ground truth reference for human annotation, and the mesh shape in our dataset comes from real-world 3D reconstruction and generation methods. We also visualize examples along with annotation scores in our dataset in \cref{fig:vis_ds}. We observe that the annotated realism score rises as the realism of a mesh shape increases.

\subsection{Evaluation Methods} 
\label{sec:eval}
To measure how closely our metric aligns with human perception, we employ three correlation coefficients. Pearson’s linear correlation coefficient (PLCC) \cite{pearson1920plcc} quantifies the linear relationship between our metric and human ratings. PLCC can be represented as:
\begin{equation}
p=\frac{\sum_{i=1}^n(\hat{s}_i-\bar{\hat{s}})(s_i-\bar{s})}{\sqrt{\sum_{i=1}^n(\hat{s}_i-\bar{\hat{s}})^2} \sqrt{\sum_{i=1}^n(s_i-\bar{s})^2}},
\end{equation}
where $\hat{s}_i$ and $s_i$ are the input mesh and ground-truth realism scores of the sample $i$, respectively, and $n$ is the data sample number. $\bar{\hat{s}}=\frac{1}{n}\sum_{i=1}^n\hat{s}_i$ and $\bar{s}=\frac{1}{n}\sum_{i=1}^n s_i$ are the mean realism prediction and ground truth realism scores.

Additionally, Spearman's ranking order correlation (SROCC) \cite{spearman1910srocc} is also used to measure the ranking order correlation between the shape realism metric and human annotation. SROCC is denoted as:
\begin{equation}
r_s=1-\frac{6\sum_{i=1}^n(R(\hat{s}_i)-R(s_i))^2}{n(n^2-1)},
\end{equation}
where \( R(\hat{s}_i) \) and \( R(s_i) \) are the rankings of \( \hat{s}_i \) and \( s_i \). \( n \) denotes. $n$ is the data sample number.

Finally, we use Kendall’s rank order correlation coefficient (KROCC) \cite{kendall1938new} as an alternative way to measure ranking order correlation. Unlike SROCC, which considers the magnitude of rank differences, this metric focuses solely on the relative ordering of ranks. KROCC can be represented as:
\begin{equation}
\tau = 1 - \frac{2}{n(n^2 - 1)} \sum_{i<j} \operatorname{sign}(\hat{s}_i - \hat{s}_j)) \operatorname{sign}(s_i - s_j),
\end{equation}
where \(\operatorname{sign}(\cdot)\) is the sign function:
\[
\operatorname{sign}(x) =
\begin{cases} 
1, & x > 0 \\ 
-1, & x < 0 \\ 
0, & x = 0 
\end{cases}
\]
Note that each coefficient ranges from -1 to 1, with higher values indicating a stronger correlation.

\begin{figure}[t]
    \centering
    \includegraphics[width=1\linewidth]{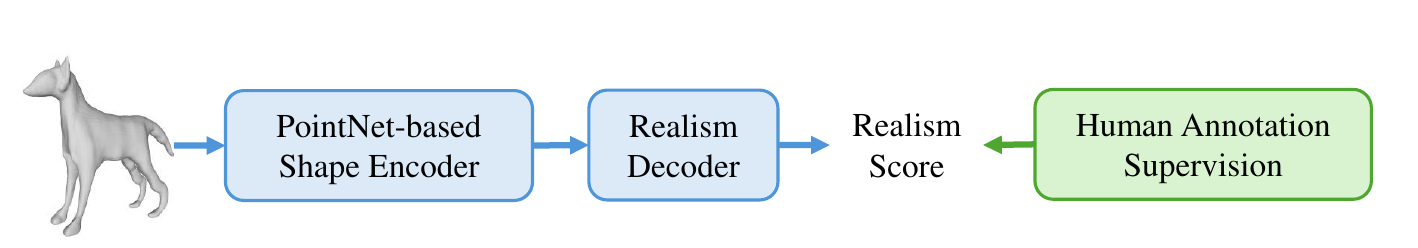}
    \caption{The point-cloud-based ad-hoc baseline design. This baseline employs PointNet \cite{qi2017pointnet} to extract shape features, which are then fed into the same realism decoder to produce a realism evaluation score.}
    \label{fig:comp}
\end{figure}

\section{Experiments}

\subsection{Implementation Details}

Following \cite{xu2024pointllm}, we adopt PointBert \cite{pointbert} as our shape encoder. The maximum number of vertices of a mesh is set to be 8192. If there are more than 8192 vertices in a mesh, we use the farthest point sampling method to select 8192 vertices. The LLM bridge is initialized with corresponding components of PointLLM \cite{xu2024pointllm}.
We adopt AdamW \cite{loshchilov2017decoupled} as our optimizer. The learning rate is set to be $2 \times 10^{-4}$. We train for 3 epochs and set the batch size to 12.

\subsection{Evaluation on \ds{} Dataset}

We evaluate the alignment between our metric and human annotations using the three evaluation methods described in Sec. \nameref{sec:eval}. The results are summarized in \cref{tab:sota}.

For comparison, we propose an ad-hoc baseline which is based on PointNet. As illustrated in \cref{fig:comp}, this baseline extracts shape features with PointNet \cite{qi2017pointnet}. The shape features are then fed into the same realism decoder to generate a realism score. 

As shown in \cref{tab:sota}, our metric demonstrates better alignment with human perception than the baseline. Our metric achieves PLCC of 0.69, SROCC of 0.70, and KROCC of 0.57. In addition, PLCC, SROCC, and KROCC of our metric are 0.13, 0.11, and 0.11 higher than those of the baseline, respectively. These numbers not only show that our metric is well aligned with human annotations, but also show that our method more faithfully reflects human perception of realism than the baseline. This illustrates the effectiveness of our LLM bridge in evaluating the realism of various 3D shapes and its generalization ability across different types of 3D shapes.

\subsection{Experiment Results}

\begin{table}[t]

\centering
\resizebox{0.8\linewidth}{!}{
    \begin{tabular}{c|c|c|c|c}
    \hline
    \multicolumn{2}{c|}{Method} & PLCC $\uparrow$ & SROCC $\uparrow$ & KROCC $\uparrow$ \tabularnewline
    \hline
    \multirow{4}{*}{LoRA} & r=2 & 0.608 & 0.626 & 0.518 \\
    & r=4 & 0.656 & 0.639 & 0.514 \\
    & r=8 & 0.550 & 0.564 & 0.441 \\
    & r=16 & 0.524 & 0.512 & 0.401 \\
    \hline
    \multicolumn{2}{c|}{Prefix Tuning} & 0.081 & 0.066 & 0.039 \\
    \hline
    \multicolumn{2}{c|}{Full Finetune (Ours)} & \textbf{0.689} & \textbf{0.696} & \textbf{0.566} \\
    \hline 
    \end{tabular}
}
\caption{Ablation Studies on finetuning methods. We compare three different methods, including two parameter-efficient finetuning methods, \textit{i.e.}, LoRA and prefix tuning, and full finetuning. We also explore how the performance changes with different LoRA ranks (denoted as ``\texttt{r}''). The results show that full fine-tuning achieves better k-fold validation results and good generalizability than the two-parameter efficient finetuning methods.} 
\label{tab:lora}
\end{table}

\textbf{Full finetuning or parameter efficient finetuning.}
We explore whether parameter-efficient finetuning methods, \textit{e.g.}, LoRA, prefix tuning, could lead to better performance. In \cref{tab:lora}, we compare
three different finetuning methods: LoRA, prefix tuning, and full finetuning. We can see from the results that prefix tuning leads to poor performance. There is very little correlation between realism scores produced by the model trained via prefix tuning and human annotations. In addition, LoRA with ranks of 2 and 4 performs better than larger ranks of 8 and 16, with rank 4 achieving the best performance PLCC (0.656) and SROCC (0.639). However, full finetuning outperforms all LoRA settings across all three metrics. These results show that while LoRA and prefix tuning offer parameter efficiency, finetuning all parameters of the LLM bridge is more beneficial for our  3D realism evaluation task.

\begin{table}

\centering
\resizebox{0.9\linewidth}{!}{
    \begin{tabular}{c|c|c|c}
    \hline
    Method & PLCC $\uparrow$ & SROCC $\uparrow$ & KROCC $\uparrow$ \tabularnewline
    \hline
    Generation & 0.245 & 0.250 & 0.217 \\
    \hline
    Regression (Ours) & \textbf{0.689} & \textbf{0.696} & \textbf{0.566} \\
    \hline 
    \end{tabular}
}
\caption{Ablation study on realism score generation methods. We compare two methods: ``\texttt{Generation}'' and ``\texttt{Regression}''. For ``\texttt{Generation}'', we train the model to predict \textit{text} tokens corresponding to quantized integer realism scores. For ``\texttt{Regression}'', we adopt the realism decoder to regress realism scores.}
\label{tab:abla}
\end{table}

\textbf{Realism score: generation or regression.}
We investigate how the realism score should be produced. Specifically, we compare two methods: ``\texttt{Generation}'' and ``\texttt{Regression}''. For ``\texttt{Generation}'', we train the model to predict \textbf{text} tokens corresponding to quantized integer realism scores (more details can be found in our supplementary materials). For
``\texttt{Regression}'', we adopt the realism decoder introduced in Sec. \nameref{sec:sram} to regress realism scores. We can see from \ref{tab:abla} that regressing the realism scores leads to much better performance. Hence, we adopt the ``\texttt{Regression}'' method.

\begin{table}[h]

\centering
\resizebox{0.9\linewidth}{!}{
    \begin{tabular}{c|c|c|c}
    \hline
    Prompt & PLCC $\uparrow$ & SROCC $\uparrow$ & KROCC $\uparrow$ \tabularnewline
    \hline
    W/ obj name & 0.567 & 0.559 & 0.443 \\
    \hline
    W/o obj name (Ours) & \textbf{0.689} & \textbf{0.696} & \textbf{0.566} \\
    \hline 
    \end{tabular}
}
\caption{Ablation study on prompt. We investigate whether including the object name in realism prompt helps our metric achieve better performance. If object name is included in the realism prompt ''\texttt{Evaluate the quality of this point cloud \textit{object} and provide your rating...}'', we replace ''\texttt{\textit{object}}'' with an object name, \textit{e.g.}, ``\texttt{dog}''. Otherwise, we keep the original realism prompt. These two prompts are denoted as ``\texttt{W/ obj name}'' and ``W/o obj name (Ours)'', respectively.}
\label{tab:prompt}
\end{table}

\textbf{Prompt: Include object name or not.}
We explore whether or not we should include the object name in the realism prompt. If object name is included in the realism prompt ``\texttt{Evaluate the quality of this point cloud object and provide your rating...}'', we replace ``\texttt{object}”
with an object name, \textit{e.g.}, ``\texttt{dog}''. Otherwise, we keep the original
realism prompt. These two prompts are denoted as ``\texttt{W/ obj name}'' and ``\texttt{W/o obj name (Ours)}'', respectively. As shown in \cref{tab:prompt}, the prompt without object name leads to better alignment of our metric with human annotations. The PLCC, SROCC, and KROCC of ``\texttt{W/o obj name}'' are at least 0.122 more than those of ``\texttt{W/ obj name}''.

\begin{figure}[t]
    \centering
    \includegraphics[trim={35pt 0 35pt 0},width=0.75\linewidth]{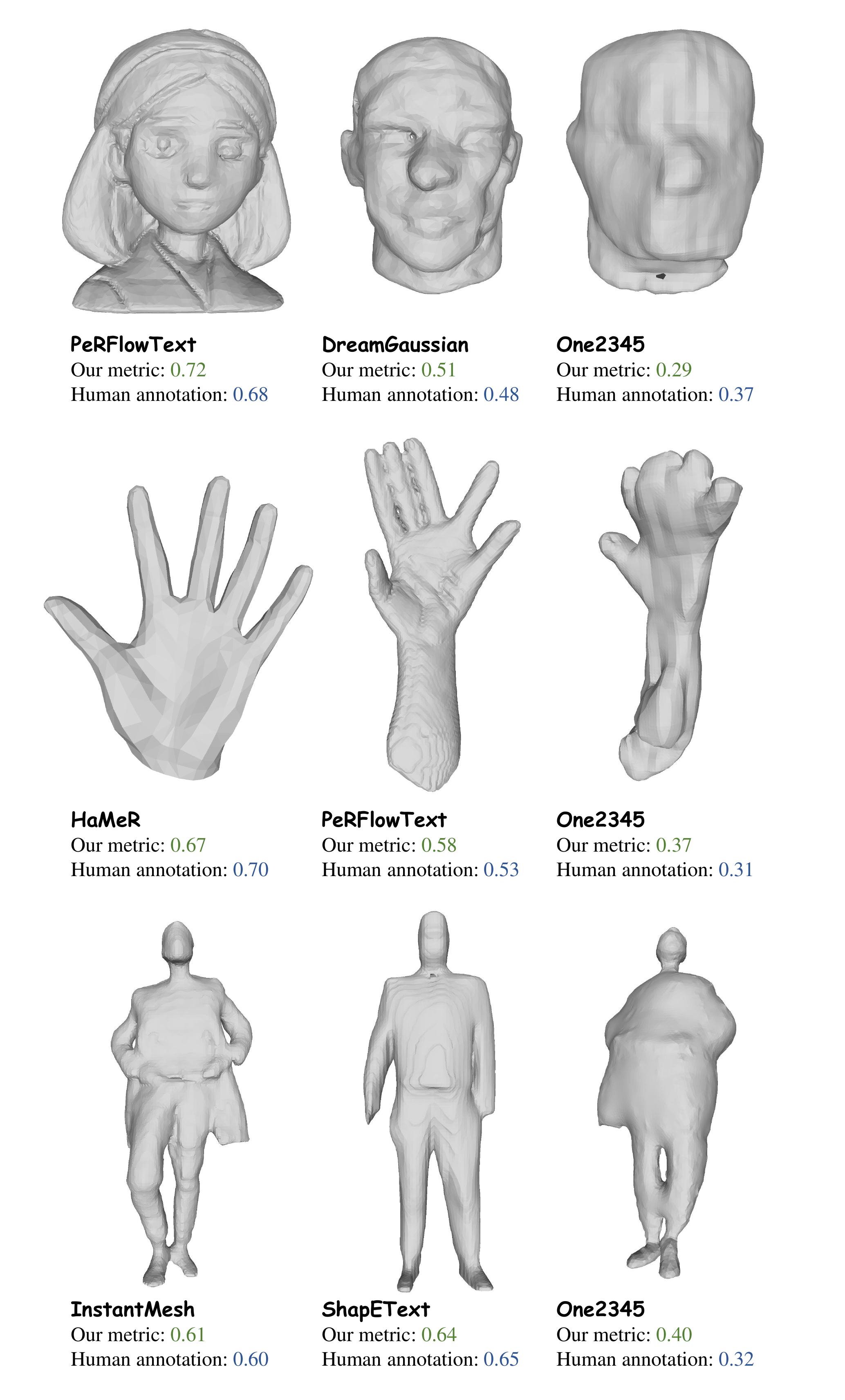}
    \caption{We present the realism scores from our metric alongside human-annotated realism scores for various meshes. The results show that our metric assigns high realism scores to realistic meshes, while severely distorted meshes receive low scores. Our metric correlates well with human annotations, which reflects how human annotators perceive mesh realism.}
    \label{fig:vis}
\end{figure}

\textbf{Visualization.}
In \cref{fig:vis}, we show some visualization of 3D shapes, realism scores of our metric, and human-annotated realism scores. We can see that our method assigns high realism scores to realistic 3D shapes, while distorted shapes receive
low scores. For example, as shown in the second row, the hand generated by HaMeR receives a score of 0.67. The hand generated by PeRFlowText received a lower score, as it has six fingers. The hand generated by One2345 receives the lowest score, as we can barely recognize that it is a hand. These visualizations illustrate that our metric correlates well with human annotations, which reflects how human annotators perceive mesh realism.

\section{Conclusion}

In conclusion, we have introduced a novel no-reference metric for evaluating 3D shape realism that operates solely on the 3D shape itself. Our method, which leverages a pretrained 3D language model and employs a LoRA-style finetuning approach, effectively integrates human realism annotations to capture high-level semantic features and align its scores with human perception. The introduction of a new dataset containing human-labeled scores for meshes generated by a diverse range of reconstruction and generation algorithms further validates the practical applicability of our approach. Experimental results demonstrate that our metric correlates strongly with human judgments and outperforms existing methods, offering a promising tool for advancing the evaluation of 3D shapes in content creation industries.

\bibliography{aaai2026}

\end{document}